\documentclass[10pt,twocolumn,letterpaper]{article}

\usepackage{wacv}
\usepackage{times}
\usepackage{epsfig}
\usepackage{graphicx}
\usepackage{amsmath}
\usepackage{amssymb}

\newcommand*{\Comb}[2]{{}^{#1}C_{#2}}%

\def\wacvPaperID{985} % Enter the WACV Paper ID here

\wacvfinalcopy % *** Uncomment this line for the final submission

\ifwacvfinal
\def\assignedStartPage{1} % *** Enter the assigned starting page number (instead of 9876)
\fi

\ifwacvfinal
\usepackage[breaklinks=true,bookmarks=false]{hyperref}
\else
\usepackage[pagebackref=true,breaklinks=true,colorlinks,bookmarks=false]{hyperref}
\fi

\ifwacvfinal
\else
\fi

\begin{document}

%%%%%%%%% TITLE
\title{Deep Active Learning with Augmentation-based Consistency Estimation}
%Deep Active Learning with Consistency-based Regularization}

\author{
SeulGi Hong\\
hutom\\
{\tt\small sghong@hutom.io}
\and
Heonjin Ha\\
hutom\\
{\tt\small hihunjin@hutom.io}

\and
Junmo Kim\\
KAIST\\
{\tt\small junmo.kim@kaist.ac.kr}

\and
Min-Kook Choi\\
hutom\\
{\tt\small mkchoi@hutom.io}
}

\maketitle
%\thispagestyle{empty}

%%%%%%%%% ABSTRACT
% 나중에 영어 고치기!
\begin{abstract}
In active learning, the focus is mainly on the selection strategy of unlabeled data for enhancing the generalization capability of the next learning cycle. For this, various uncertainty measurement methods have been proposed. On the other hand, with the advent of data augmentation metrics as the regularizer on general deep learning, we notice that there can be a mutual influence between the method of unlabeled data selection and the data augmentation-based regularization techniques in active learning scenarios. Through various experiments, we confirmed that consistency-based regularization from analytical learning theory could affect the generalization capability of the classifier in combination with the existing uncertainty measurement method. By this fact, we propose a methodology to improve generalization ability, by applying data augmentation-based techniques to an active learning scenario. For the data augmentation-based regularization loss, we redefined cutout (co) and cutmix (cm) strategies as quantitative metrics and applied at both model training and unlabeled data selection steps. We have shown that the augmentation-based regularizer can lead to improved performance on the training step of active learning, while that same approach can be effectively combined with the uncertainty measurement metrics proposed so far. We used datasets such as FashionMNIST, CIFAR10, CIFAR100, and STL10 to verify the performance of the proposed active learning technique for multiple image classification tasks. Our experiments show consistent performance gains for each dataset and budget scenario. The source code will open to the public.
\end{abstract}

%\vspace{-4mm}     %
%%%%%%%%% BODY TEXT
\section{Introduction}

Active learning is a method that estimates the uncertainty of unlabeled data to select candidates to be labeled which can improve the learning curve on a given budget scenario. Beyond image recognition, it has been widely studied to improve the learning efficiency in medical image recognition or semantic segmentation problems, where the cost of labels is very high \cite{Settles10}. Recently, as the deep learning based architectures have shown excellent performance in many fields and there is more access to libraries for deep learning, research is actively conducted to apply a deep neural network to active learning scenarios \cite{Gal17,Beluch18,Sener18,Hu19,Yoo19}.

However, in active learning research using deep neural networks, 
%the classical and well-known approach of selecting data to be labeled based on an uncertainty measurement technique, is mostly used. In this classical approach, 
the role of deep neural networks does not actively reflect the \emph{inherent characteristics of deep neural network training methods} in active learning scenarios, except for the role of feature encoders that provide learning representation \cite{Gal17,Beluch18,Sener18}. Recently, active learning research using structural features of deep neural networks has been proposed \cite{Yoo19}. In \cite{Yoo19}, a submodule of a convolutional neural network (CNN) was used to estimate the expected loss in the process of training the learning representation for the target task, assuming uncertainty for the unlabeled data during the CNN training. However, various methodologies that utilize the characteristics of deep neural network training for active learning scenarios still need to be studied.

\begin{figure*}[t!]
\centering
\includegraphics[width=1.0\linewidth]{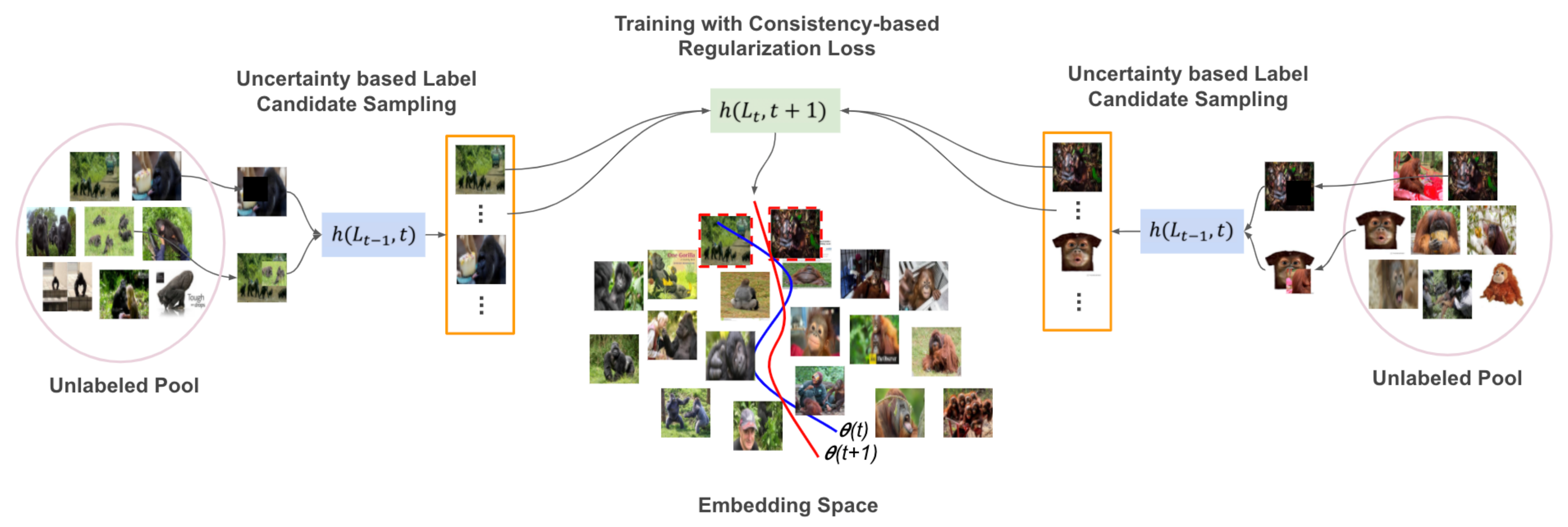}
%\vspace{-5mm}
\caption{\textbf{A schematic depiction of the proposed active learning method.} $h(L_{t-1},t)$ is a model trained with labeled data $L_{t-1}$ in $t$ cycles in an active learning scenario, and $h(L_t,t+1)$ is a model trained with labeled data $L_t$ in $t+1$ cycle. $\theta(t)$ represents the decision boundary generated in the embedding space for classification from the model $h(L_{t-1},t)$ and $\theta(t+1)$ represents the decision boundary generated from the model $h(L_t,t+1)$. In the embedding space, the samples indicated by the red dotted lines are the samples selected as the samples to be labeled in the $t+1$ cycle with high uncertainties among the unlabeled samples of each class. In this case, the decision boundary generated in each cycle may be influenced by the method of sample selection and the training strategy.}
\label{fig:example}
%\vspace{-5mm}
\end{figure*}

Furthermore, there is an unnoticed point in active learning that researches show improved performance only in \emph{specific datasets and budget scenarios}. \cite{Simeoni19} showed that methods still have poor generalization performance when datasets or budget conditions change. To overcome this problem, our approach is motivated by the analytical learning theory \cite{Kawaguchi18}. In short, analytical learning theory suggests a way to measure the generalization gap approximately, and \cite{Kawaguchi18} also shows how the theory can be applied to deep learning as a consistency-based regularizer. We adopt the idea of the cutout and cutmix data augmentation techniques \cite{DeVries17,Yun19} and we devised consistency-based regularization losses and uncertainty measures. Our suggesting methods pursue to improve the generalization capabilities of deep neural network models, to actively reflect the characteristics of deep neural network training in active learning. Inspired by existing research \cite{Kawaguchi18}, we propose a learning methodology that maximizes efficiency which is insensitive to budgets and datasets by combining uncertainty measurement techniques and regularization loss for each step of active learning scenarios: candidate sampling step and training step.
%At the same time, there is a need for improvement as studies related to active learning using deep neural networks show improved performance only in specific datasets and budget scenarios, but poor learning performance when datasets or budget scenarios change \cite{Simeoni19}. We use data augmentation-based regularization techniques \cite{DeVries17,Yun19} and consistency-based regularization losses using analytical learning theory to improve the generalization capabilities of CNNs, to actively reflect the characteristics of deep neural network training in active learning. Inspired by existing research \cite{Kawaguchi18}, we propose a learning methodology that maximizes training efficiency while being insensitive to budgets and datasets by combining uncertainty measurement techniques and regularization loss in active learning scenarios.

Besides, various image transformation techniques are applied to mini-batches for robustness during the learning process of the image recognition models \cite{Krizhevsky12,Simonyan15,He16}. Afterward, data augmentation-based regularization techniques \cite{DeVries17,Yun19,Zhang18}, which are derived from the image transformation, have been proposed as a major factor in deep neural network training to induce direct generalization performance improvement of CNNs \cite{Hernandez-Garcia19}. We attempted to reflect the characteristics of these data augmentation techniques to active learning, which are generally used in the training of deep neural network. To do this, we first observed that data augmentation-based regularization techniques could be a major factor in measuring the generalization capability of unlabeled data. In order to use this observation, we made use of analytical learning theory to estimate the uncertainty, by using the measurement metric of the variation of function based on the generalized performance boundary approximation \cite{Kawaguchi18}. A schematic depiction of our active learning technique is shown in Figure 1. 
%In the case of CNN training for image recognition, methods for training robust classifiers against noise and change by applying various image transformation techniques to mini batches in the learning process have been proposed \cite{Krizhevsky12,Simonyan15,He16}.

%----- 정의 추가--- emph를 빼고 따옴표를 치는 게 낫나 고민
Our suggesting methods are basically derived from the augmentation-based consistency estimation. For the training step of active learning, we devised \emph{augmentation-based consistency measurement (estimation) for regularization loss}. We will call it as consistency-based regularization or augmentation-based regularization for convenience.  For the candidate sampling step of active learning, we suggest \emph{augmentation-based uncertainty measurement}, briefly saying augmentation-based uncertainty or uncertainty measure.

%----- 정의 끝---

The technical contributions of the active learning method using the consistency-based regularization are as follows.
%\vspace{-1mm}
\begin{itemize}
  \item We apply the augmentation-based measurement metric for \emph{the candidate sampling step} of active learning scenarios as an uncertainty measure. To naturally apply augmentation-based techniques to active learning, we propose a method of measuring the variation of function derived from the analytical learning theory. By doing so, our algorithm selects the data which is hard for the current model to ensure consistency.
  \item Consistency-based regularization can be adapted to \emph{the training step} and contributed to active learning as a regularizer. In particular, the regularization loss combined with any uncertainty measurement (existing and our suggesting) techniques yielded an improved generalization performance.
  \item Through \emph{deep active learning} with our suggesting consistency-based methods, we show the steady improvement of generalization performance that is not significantly affected by dataset and budget scenario changes in the image classification.
\end{itemize}

%-------------------------- BACKGROUND ----------------------------------------------
\section{Background}
\noindent \textbf{Active Learning.}  Active learning in machine learning has been proposed mainly as a method of selecting the samples to be labeled by measuring the uncertainty of the unlabeled samples. Representative uncertainty measurement methods use the entropy for the probability distribution of the classifier as a processed input feature \cite{Holub08} or the margin between the input feature and the hypothesis function \cite{Balcan07}, or the margin between the feature vectors \cite{Sener18}. These uncertainty measurement methods are also used for active learning using CNNs, resulting in improved performance in image recognition problems. \cite{Kading18} used Gaussian process models for expected model output changes for active learning, and \cite{Gal17} applied Bayesian inference based on dropout to use probabilistic approaches using Bayesian prior to active learning. Efforts have also been made to improve the efficiency of active learning by using ensemble approaches \cite{Beluch18} to apply a Query-by-Committee strategy to deep neural networks or by predicting expected losses with submodules that rely on recognition modules for target tasks \cite{Yoo19}.

However, many active learning techniques in the image recognition field using deep neural networks show performance sensitivity to the dataset and budget scenarios, and it has been reported that the actual performance is difficult to reproduce \cite{Simeoni19}. Active learning, using our proposed consistency-based regularization, is more deeply considered in the learning methodology for deep neural networks than the previous active learning techniques, and provides consistent performance improvements that are independent of the dataset and budget scenarios. \\

%\vspace{-2mm}
\noindent \textbf{Data Augmentation-based Regularization.} Data augmentation-based regularization techniques help to improve generalization performance in deep neural networks training. Unlike the explicit regularization methods such as weight decay which is applied with a statistical learning approach, data augmentation-based regularization is aimed at preventing overfitting through transformation on input data during training \cite{Zhang17}. Data augmentation-based regularization techniques are applied mainly using transformation methods but recently proposed data augmentation-based regularization techniques \cite{DeVries17,Yun19,Zhang18} have used intentional data mixing and soft labeling \cite{Hinton14} to improve the generalization performance. 
Active learning, which requires efficient learning using such a limited amount of training data, is inextricably linked with data augmentation-based regularization. We reformatted both the cutout \cite{DeVries17} and cutmix \cite{Yun19} among the data augmentation-based regularization techniques to assist in the active learning cycle. Using redefined data augmentation-based regularization, we could apply both the process of selecting the data to be labeled and the process of training the classifier in each active learning cycle. \\

%여기 ALT 설명 틀릴 수도 있어서 주의
\noindent \textbf{Analytic Learning Theory.} Analytical learning theory is a measure-theoretic learning approach for machine learning suggested in \cite{Kawaguchi18}. It is based on a non-statistical method that minimizes model assumptions about the data and uses the characteristics of the data according to empirical observations. Because model training using deep neural networks often does not follow known probability distributions or model assumptions, various analyses based on empirical observations have been reported \cite{Zhang17}.
Our proposed active learning method is based on \cite{Kawaguchi18} that transforms a data-based regularization method into a consistency-based loss by analyzing the generalized error bound of deep neural networks through the analysis learning theory. 
We extend the measurement of the variation of function derived from the analytical learning theory from dual-cutout \cite{Kawaguchi18} to k-cutout and apply it to each active learning cycle. To adopt cutmix augmentation for active learning, we add some constraints on the definition of cutmix strategy to preserve the data semantics. For tne training cycle of deep active learning, we redefine the cutmix method using soft labels into the consistency-based regularization loss according to the analytical learning theory. In addition, we convert cutmix regularization into the entropy-based uncertainty measurement technique for the candidate sampling step of active learning. By redefining the data augmentation-based techniques, we can apply them to active learning scenarios and develop training strategies that are less affected by dataset and budget constraints.
% 이게 수정 전 원본
%\noindent \textbf{Analytic Learning Theory.} Analytical learning theory is based on a non-statistical measurement theory that minimizes model assumptions about the data and uses the characteristics of the data according to empirical observations. Because model training using deep neural networks often does not follow known probability distributions or model assumptions, various analyses based on empirical observations have been reported \cite{Zhang17}. The proposed active learning method is based on the study that transforms a data-based regularization method into a consistency-based loss by analyzing the generalized error bound of deep neural networks through the analysis learning theory \cite{Kawaguchi18}. We extend the measurement of the variation of function derived from the analytical learning theory from dual-cutout \cite{Kawaguchi18} to k-cutout, while converting the cutmix method using soft labels into the entropy-based uncertainty measurement technique and applying it to the consistency-based loss. By redefining the data augmentation-based regularization techniques, we can apply them to active learning scenarios and develop training strategies that are less affected by dataset and budget constraints.

%--------------------------- METHOD ---------------------------------------------
% 3.2에 직관적인 설명이 부족하다는 코멘트
% 원래 말하고자 했던 문장이랑 다르게 번역된 문장 (수정하였음)
%\vspace{-3mm}
\section{Deep Active Learning with Consistency-based Regularization}
%\vspace{-1mm}
This section describes active learning using our proposed consistency-based techniques. First, we define problems and notations for active learning scenarios and then explain how the estimation of variation of function in active learning scenarios is related to the generalization of deep neural networks. We also describe an approach to apply cutout and cutmix as uncertainty measures, using representative data augmentation, to active learning scenarios. In addition, we provide a semantic visualization of how consistency-based loss with data augmentation affects the total loss function during training, thus explaining why the proposed training strategy shows improved performance.

\subsection{Background and problem formulation}
%\vspace{-2mm}
To define an active learning scenario in a $t$ cycle for sample set $S=\{(x_1,y_1), (x_2,y_2),$ $... ,(x_N,y_N)\}$ for data and label pairs, it consists of labeled data $\mathcal{L}_{b,t}$ and unlabeled data $\mathcal{U}_{N-b,t}$ in the entire dataset $\mathcal{D}_N=\mathcal{L}_{b,t}+\mathcal{U}_{N-b,t}$. At this time, the initial training samples are fixed to $b$ uniform samples $\mathcal{L}_{b,0}$. Active learning consists of $b(t+1)$ labeled data and $N-b(t+1)$ unlabeled data in $\mathcal{D}_N$ according to cycle $t$ and budget $b$. In this case, the active learning process is performed by selecting $b$ data to be labeled in the next cycle among the unlabeled data $\mathcal{U}_{N-b,t}$ in every training cycle. Thus, when selecting $b$ data to be labeled in a particular cycle $t$ of active learning, the following conditions must be met for the classifier $h(x,y;\theta)$ to be trained:

\setlength{\belowdisplayskip}{0pt} \setlength{\belowdisplayshortskip}{0pt}
\setlength{\abovedisplayskip}{-2pt} \setlength{\abovedisplayshortskip}{-2pt}
%\vspace{-1mm}
% equation1
\begin{align}
  \operatorname*{argmax}_{\mathcal{L}_{b,t+1}-\mathcal{L}_{b,t}, \theta_{t+1}} E[ h_{\mathcal{L}_{b,t+1}}(x,y;\theta_{t+1})] - E[ h_{\mathcal{L}_{b,t}}(x,y;\theta_{t}) ],
\end{align}

\noindent where $E$ is the generalization error of classifier $h$ and $h_{\mathcal{L}_{b,t}}(x,y;\theta_{t})$ is a classifier with parameter $\theta_t$ trained in $t$ cycles using labeled data $\mathcal{L}_{b,t}$. According to Equation (1), active learning is a problem of finding the subset $\mathcal{L}_{b,t+1}-\mathcal{L}_{b,t}\subseteq \mathcal{U}_{N-b,t}$ of the data having the largest generalization gap between the previous cycle and the current cycle and the classifier parameter $\theta_{t+1}$. To satisfy this condition, we must define a selection strategy for finding good $\mathcal{L}_{b,t+1}-\mathcal{L}_{b,t}$ and a learning strategy for finding good $\theta_{t+1}$ at cycle $t+1$.

% 이부분 직관적인 설명 필요. (했었는데 리뷰어가 지적)
\subsection{Variation of function estimation via analytical learning theory}

We used an approach that utilizes the analytical learning theory for the generalized error bound in machine learning proposed in \cite{Kawaguchi18}, to simultaneously achieve two active learning goals given in Equation (1). The generalized error bound based on the analytical learning theory demonstrated in \cite{Kawaguchi18} is given by:

%\vspace{-1mm}
% equation2
\begin{align}
\begin{split}
  &E[ h_{\mathcal{L}}(x,y;\theta)] - \hat{E}[ h_{\hat{\mathcal{L}}}(x,y;\hat{\theta}) ] \\\leq 
  &\sum_{y\in\mathcal{Y}}c_2p(y)\sigma(f_y)\sqrt{\frac{d_z}{|\hat{\mathcal{L}}_{x|y}|}} +
  \hat{E}[ h_{\hat{\mathcal{L}}}(x,y;\hat{\theta}) ] \sqrt{\frac{\log(2/\delta)}{2|\hat{\mathcal{L}}_{x|y}|}},
  \end{split}
\end{align}
%\vspace{-1mm}

\noindent where $\mathcal{L}$ is a theoretical pair of labeled data that can minimize generalization errors and $\hat{\mathcal{L}}$ is a given set of labeled data for actual training. $E[ h_{\mathcal{L}}(x,y;\theta)] - \hat{E}[ h_{\hat{\mathcal{L}}}(x,y;\hat{\theta}) ]$ is a generalization gap for $\mathcal{L}$ and the labeled subset $|\hat{\mathcal{L}}_{x|y}|$ is given by $|\hat{\mathcal{L}}_{x|y}|\subseteq\hat{\mathcal{L}}$, $p(y)\triangleq\frac{|\hat{\mathcal{L}}_{x|y}|}{|\hat{\mathcal{L}}|}$. In this case, $d_z$ given in the dimension of the $z$-level hidden layer and the constant $c_2$ are developed in Proposition 2 of \cite{Kawaguchi18}. The adjustable term for the generalized error boundary during training is the amount of $\sigma(f_y)$ variation of function. According to Equation (2), minimizing $\sigma(f_y)$ minimize the upper boundary of generalization error. \cite{Kawaguchi18} proposed a consistency-based regularization loss using dual-cutout augmentation on input data \cite{DeVries17} to reflect $\sigma(f_y)$ during training and is defined as:

%\vspace{-1mm}
% equation3
\begin{align}
\begin{split}
  &L_{reg}(x,\theta)=\\
  &\int_{(x^{co}_1,x^{co}_2)} \parallel h(x^{co}_1;\theta)-h(x^{co}_2;\theta) \parallel^2_2dP(x^{co}_1,x^{co}_2|x),
\end{split}
\end{align}

\noindent where $P(x^{co}_1,x^{co}_2|x)$ is defined as two random cutouts for one input data. We have modified and redefined the cutout and cutmix to fit the active learning scenario using the definition of consistency-based regularization loss in Equation (3). At the same time, the regularization technique using data augmentation was scored and applied to the active learning cycle with uncertainty for the selection of data to be learned in the next cycle.

% 이 문단은 그다지 고칠 것 X. 내용상으로도 틀린 것 없음
\subsection{Deep active learning with $k$-cutout}

First, we extend the random cutout generation from dual to $k$ in order to use cutout for the uncertainty measurement method of dual cutout proposed in \cite{Kawaguchi18}. The $k$-cutout for uncertainty measurement is then defined as:

%\vspace{-1mm}
% equation4
\begin{align}
\begin{split}
  &U(x,\theta)=\\
  &\int_{(x^{co}_i,x^{co}_j)} \parallel h(x^{co}_i;\theta)-h(x^{co}_j;\theta) \parallel^2_2 dP(x^{co}_i,x^{co}_j|x)\\
  &\approx \frac{1}{\Comb{K_{co}}{2}} \sum^{\Comb{K_{co}}{2}}_{1} (h(x^{co}_i;\theta_t)-h(x^{co}_j;\theta_t))^2, (i \neq j),\\
\end{split}
\end{align}
%\vspace{-1mm}

\noindent where $\Comb{K_{co}}{2}$ is the number of pairwise cases when $K_{co}$ random cutout images are generated and $x^{co}_i$ is input data applying an arbitrary cutout to the $i^{th}$ sample. $K_{co}$ random cutout images were generated for $k$-cutout and the mean value of the pairwise distance for each inference result was defined as uncertainty. Based on the estimated uncertainty, we trained the data with the large amount of variation of function first and expected the effect of minimizing the generalized error boundary to a greater extent than other unlabeled data. At the same time, it was expected to show good generalization performance for training after data selection in combination with consistency-based regularization loss. Consistency-based regularization loss using cutout when training about the current cycle is defined as follows.

%\vspace{-1mm}
% equation5
\begin{align}
\begin{split}
  &L_{co}(x,\theta)=\\
  &\frac{1}{M}(\frac{1}{\Comb{K_{co}}{2}}\sum^{\Comb{K_{co}}{2}}_{1} (h(x^{co}_i;\theta_{t+1})-h(x^{co}_j;\theta_{t+1}))^2 \\
  &+ \frac{1}{K_{co}}\sum^{K_{co}}_{1}CE(h(x^{co}_{i};\theta_{t+1}),y_i)), (i \neq j),
\end{split}
\end{align}
%\vspace{-1mm}

\noindent where $M$ represents the size of the mini-batch and $CE$ represents the cross-entropy function. In Equation (5), two regularization terms are applied to efficiently reflect the variation of function through cutout during training. The first term is a regularization term for minimizing mean squared error (MSE) for different $k$-cutout samples. The network must be trained in such a direction as to minimize the variation of function for $K_{co}$ number of cutout samples. The second term is the cross-entropy output with the ground truth label for the $k$-cutout samples, which includes a condition to minimize the amount of variation of function and deduce the correct answer with the corrupted image. The total loss using $k$-cutout is given by $L_{total}=L_{ce}+L_{co}$, where $L_{ce}$ is cross-entropy loss for the target task.

% FIGURE2
 \begin{figure*}[t!]
%\vspace{-5mm}
\centering
\includegraphics[width=1.0\linewidth]{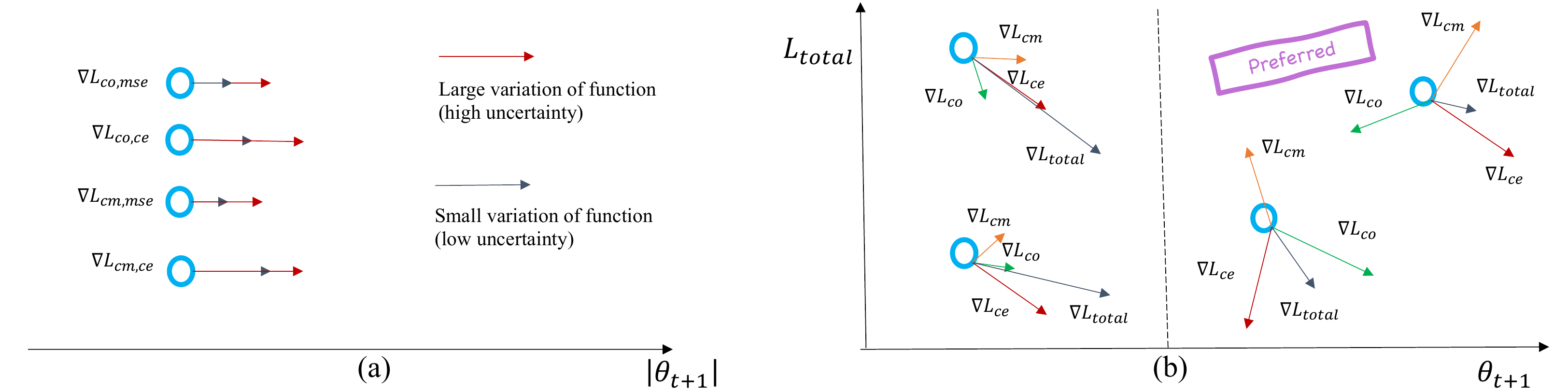}
%\vspace{-4mm}
\caption{\textbf{Semantic representation of the effects of uncertainty and variation of function on model training.} (a) shows the tendency of the magnitude of gradient that the cutmix and cutout get into the loss function. (b) shows the effect of two samples on total loss due to sample uncertainty.}
\label{fig:example}
%\vspace{-5mm}
\end{figure*}

% 영어 어순
%\vspace{-3mm}
\subsection{Deep active learning with cutmix}
%\vspace{-1mm}
In addition to the cutout, we applied the cutmix technique \cite{Yun19} to measure uncertainty and loss of consistency-based regularization for active learning scenarios, since cutmix is known to improve the generalization performance by data augmentation-based regularizations and soft label. To measure the uncertainty based on cutmix, we need $K_{cm}$ different cutmix processes on one unlabeled sample and calculate the entropy of inference output $h(x_i;\theta_t)$. At this time, we choose an unlabeled sample that shows unstable inference results for several cutmix images with different semantics, expecting that it can stabilize the variation of the function of the $h(x_i;\theta_t)$ by adding it into the labeled set.

%\vspace{-1mm}
% equation6
\begin{align}
  U(x,\theta)=-\frac{1}{K_{cm}}\sum^{K_{cm}}_{1}h(x_i;\theta_t)\log h(x_i;\theta_t).
\end{align}
%\vspace{-1mm}

Meanwhile, the consistency-based regularization loss with cutmix is defined as follows:

% equation7
\begin{align}
\begin{split}
  &L_{cm}(x,\theta)=\\
  &\frac{1}{M}(\frac{1}{\Comb{K_{cm}}{2}}\sum^{\Comb{K_{cm}}{2}}_{1} (h(x^{cm}_i;\theta_{t+1})-h(x^{cm}_j;\theta_{t+1}))^2 \\
  &+ \frac{1}{K_{cm}}\sum^{K_{cm}}_{1}CE(h(x^{cm}_{i};\theta_{t+1}),y_i)), (i \neq j),
\end{split}
\end{align}
\vspace{-1mm}
%\begin{align}
%  L_{cm}(x,\theta)=\frac{1}{M}(\frac{1}{\Comb{K_{cm}}{2}}\sum^{\Comb{K_{cm}}{2}}_{1} (h(x^{cm}_i;\theta_{t+1})-h(x^{cm}_j;\theta_{t+1})) ^2
%  \notag\\+ \frac{1}{K_{cm}}\sum^{K_{cm}}_{1}CE(h(x^{cm}_{i};\theta_{t+1}),y_i)), (i \neq j),
%\end{align}

\noindent where $\Comb{K_{cm}}{2}$ is the number of pairwise cases for $K_{cm}$ arbitrary cutmix images of the $i^{th}$ sample. The consistency-based regularization loss using cutmix is defined as the linear combination of the distance metric and the cross-entropy output, in the same way as the cutout-based regularization loss. The reason for not including the entropy term used to measure uncertainty in the cutmix-based regularization loss, is to avoid using additional balance parameters or normalization constants for the total loss. In this case, the total loss is defined as $L_{total}=L_{ce}+L_{cm}$. Finally, the total loss of both cutout and cutmix is given by $L_{total}=L_{ce}+L_{co}+L_{cm}$. The gradient of the backpropagation for the total loss during training in $t+1$ cycle is then obtained by $\nabla L_{total}=\nabla L_{ce}+\nabla L_{co}+\nabla L_{cm}$.

% 문장을 최대한 풀어 쓰려고는 했는데, 문장이 읽기 편한지가 의문
The semantic visualization of the backpropagation process in Figure 2 shows the role of our consistency-based regularization losses. It depends on the derivative of the total loss.
In Figure 2(a), the length of each arrow indicates the \emph{magnitude} of the gradient. It means how much each loss could affect updating model parameters.
In the case of cutmix, the cross-entropy loss $L_{cm,ce}=\frac{1}{K_{cm}}\sum^{K_{cm}}_{1}CE(h(x^{cm}_{i};\theta_{t+1})$ is derived from differences comparing with correct labels. It has a larger magnitude of gradient than MSE loss  $L_{cm,mse}=\frac{1}{\Comb{K_{cm}}{2}}\sum^{\Comb{K_{cm}}{2}}_{1}$ $(h(x^{cm}_i;\theta_{t+1})-h(x^{cm}_j;\theta_{t+1}))^2$, and it is occurred from differences between the inference output of augmented data from the same sample.
The red arrow shows an example of the gradient \emph{magnitude} that occurred in the sample with large uncertainty, and the black one indicates another sample with small uncertainty.

% TABLE1
\begin{table*}[t!]
  \caption{\textbf{Active learning results using the FashionMNIST dataset ($b=300$).} Each cell represents the performance of a cycle $t$ for each active learning technique, which is measured 3 times. Values shown in \textcolor{red}{red} indicate improved accuracy compared to random samples, while values in bold \textbf{\textcolor{blue}{blue}} indicate the highest accuracy of any active learning strategy. The last row shows the difference in accuracy between random samples and the active learning strategy with the greatest performance gains.
}
  \label{tab:freq}
  \resizebox{\linewidth}{!}{
  \begin{tabular}{c|c|c|c|c|c|c|c|c|c|c|c}
%    \toprule
\hline
   uncertainty/loss & 0 & 1 & 2 & 3 & 4 & 5 & 6 & 7 & 8 & 9 & 10\\ \hline
 %   \midrule
	random/task & 63.51 & $77.84\pm0.95$ & $80.07\pm0.41$ & $83.91\pm0.47$ & $84.55\pm0.98$ & $86.12\pm0.19$ & $86.99\pm0.34$ & $86.66\pm0.32$ & $87.61\pm0.66$ & $88.65\pm0.38$ & $89.15\pm0.13$\\ 
	cutout/task & 63.51 & $74.29\pm0.49$ & $79.01\pm0.41$ & $83.26\pm0.17$ & \textcolor{red}{$85.1\pm0.1$} & $86.09\pm0.25$ & $86.94\pm0.13$ & \textcolor{red}{$87.53\pm0.13$} & $87.06\pm0.29$ & $87.62\pm0.06$ & $88.19\pm0.13$\\
	cutmix/task & 63.51 & {$73.61\pm2.5$} & \textcolor{red}{$81.82\pm0.58$} & {$82.07\pm1.44$} & {$83.35\pm0.53$} & \textcolor{red}{$86.12\pm0.08$} & {$86.63\pm0.59$} & \textcolor{red}{$87.17\pm0.67$} & {$87.53\pm0.54$} & {$88.35\pm0.41$} & {$88.49\pm0.82$}\\ 
	entropy/task & 63.51 & {$76.86\pm0.26$} & \textcolor{red}{$81.96\pm0.27$} & \textcolor{red}{$85.36\pm0.47$} & \textcolor{red}{$86.34\pm0.48$} & \textcolor{red}{$87.74\pm0.66$} & \textcolor{red}{$88.66\pm0.4$} & \textcolor{red}{$89.07\pm0.2$} & \textcolor{red}{$89.7\pm0.2$} & \textcolor{red}{$90.63\pm0.16$} & \textcolor{red}{$91.01 \pm 0.01$}\\ 
	margin/task & 63.51 & {$77.53\pm0.36$} & \textcolor{red}{$83.0\pm0.66$} &  \textcolor{red}{$86.5\pm0.37$} & \textcolor{red}{$87.95\pm0.29$} & \textcolor{red}{$89.11\pm0.09$} & \textcolor{red}{$90.1\pm0.34$} & \textcolor{red}{$90.42\pm0.14$} & \textcolor{red}{$90.49\pm0.17$} & 
	\textcolor{red}{$91.29\pm0.25$} & \textcolor{red}{$91.51\pm0.03$}\\ \hline
	cutout/task+co & 63.51 & \textcolor{red}{$79.71\pm0.91$} & \textcolor{red}{$84.81\pm0.4$} & \textcolor{red}{$85.75\pm0.58$} & \textcolor{red}{$87.46\pm0.73$} & \textcolor{red}{$88.9\pm0.56$} & \textcolor{red}{$89.21\pm0.52$} & \textcolor{red}{$89.23\pm0.03$} & \textcolor{red}{$89.75\pm0.07$} & \textcolor{red}{$90.09\pm0.11$} & \textcolor{red}{$89.97\pm0.03$}\\ 
	cutmix/task+co & 63.51 & \textbf{\textcolor{blue}{$81.5\pm1.17$}} & \textcolor{red}{$85.25\pm0.74$} & \textbf{\textcolor{blue}{$87.82\pm0.5$}} & \textcolor{red}{$88.38\pm0.21$} & \textcolor{red}{$89.32\pm0.5$} & \textcolor{red}{$90.0\pm0.14$} & \textcolor{red}{$90.08\pm0.36$} & \textcolor{red}{$90.64\pm0.56$} & \textcolor{red}{$90.85\pm0.12$} & \textcolor{red}{$91.33\pm0.15$}\\ \hline
	cutout/task+cm & 63.51 & \textcolor{red}{$80.06\pm0.34$} & \textcolor{red}{$83.66\pm0.25$} & \textcolor{red}{$86.4\pm0.25$} & \textcolor{red}{$87.38\pm0.38$} & \textcolor{red}{$88.4\pm0.32$} & \textcolor{red}{$89.0\pm0.19$} & \textcolor{red}{$89.1\pm0.08$} & \textcolor{red}{$89.4\pm0.38$} & \textcolor{red}{$89.84\pm0.29$} & \textcolor{red}{$89.89\pm0.3$} \\ 
	cutmix/task+cm & 63.51 & \textcolor{red}{$80.19\pm0.85$} & \textbf{\textcolor{blue}{$85.98\pm0.69$}} & \textcolor{red}{$87.02\pm0.56$} & \textcolor{red}{$88.18\pm0.54$} & \textcolor{red}{$89.59\pm0.41$} & \textcolor{red}{$89.28\pm0.21$} & \textcolor{red}{$90.44\pm0.29$} & \textcolor{red}{$90.59\pm0.31$} & \textcolor{red}{$91.12\pm0.13$} & \textcolor{red}{$91.25\pm0.05$} \\\hline
	cutout/task+co+cm & 63.51 & \textcolor{red}{$80.4\pm0.74$} & \textcolor{red}{$85.0\pm0.36$} & \textcolor{red}{$86.91\pm0.44$} & \textcolor{red}{$87.67\pm0.21$} & \textcolor{red}{$89.44\pm0.37$} & \textcolor{red}{$89.55\pm0.26$} & \textcolor{red}{$89.42\pm0.25$} & \textcolor{red}{$90.11\pm0.11$} & \textcolor{red}{$90.45\pm0.11$} & \textcolor{red}{$90.19\pm0.4$}\\
	cutmix/task+co+cm & 63.51 & \textcolor{red}{$81.02\pm0.73$} & \textcolor{red}{$85.96\pm0.52$} & \textcolor{red}{$87.79\pm0.79$} & \textbf{\textcolor{blue}{$88.9\pm0.62$}} & \textbf{\textcolor{blue}{$90.08\pm0.41$}} & \textbf{\textcolor{blue}{$90.54\pm0.5$}} & \textbf{\textcolor{blue}{$91.19\pm0.3$}} & \textbf{\textcolor{blue}{$91.55\pm0.46$}} & \textbf{\textcolor{blue}{$91.47\pm0.12$}} & \textbf{\textcolor{blue}{$91.9\pm0.15$}}\\ \hline
	maximum gap & 0 & 3.66 & \textbf{5.91} & 3.91 & 4.35 & 3.96 & 3.55 & 4.53 & 3.94 & 2.82 & 2.75\\
    \hline
    \end{tabular}
}
%\vspace{-3mm}
\end{table*}

% TABLE2
\begin{table*}[t!]
  \caption{\textbf{Active learning results using consistency-based regularization loss in the FashionMNIST dataset ($b=300$).} Combined with various uncertainty measures, the results show improved performance.
}
  \label{tab:freq}
  \resizebox{\linewidth}{!}{
  \begin{tabular}{c|c|c|c|c|c|c|c|c|c|c|c}
%    \toprule
\hline
   uncertainty/loss & 0 & 1 & 2 & 3 & 4 & 5 & 6 & 7 & 8 & 9 & 10\\ \hline
 %   \midrule
	random/task & 63.51 & $77.84\pm0.95$ & $80.07\pm0.41$ & $83.91\pm0.47$ & $84.55\pm0.98$ & $86.12\pm0.19$ & $86.99\pm0.34$ & $86.66\pm0.32$ & $87.61\pm0.66$ & $88.65\pm0.38$ & $89.15\pm0.13$ \\ \hline
	entropy/task+co & 63.51 & \textcolor{red}{$81.93\pm0.36$} & \textcolor{red}{$85.44\pm0.45$} & \textcolor{red}{$86.4\pm1.15$} & \textcolor{red}{$87.82\pm0.63$} & \textcolor{red}{$89.13\pm0.3$} & \textcolor{red}{$89.85\pm0.39$} & \textcolor{red}{$89.93\pm0.21$} & \textcolor{red}{$90.74\pm0.3$} & \textcolor{red}{$91.17\pm0.29$} & \textcolor{red}{$91.32\pm0.05$}\\
	margin/task+co & 63.51 & \textcolor{red}{$83.6\pm0.61$} & \textcolor{red}{$87.53\pm0.52$} & \textcolor{red}{$88.69\pm0.55$} & \textbf{\textcolor{blue}{$90.09\pm0.41$}} & \textbf{\textcolor{blue}{$91.44\pm0.19$}} & \textcolor{red}{$91.64\pm0.15$} & \textbf{\textcolor{blue}{$92.09\pm0.13$}} & \textcolor{red}{$92.08\pm0.2$} & \textcolor{red}{$92.24\pm0.37$} & \textcolor{red}{$92.56\pm0.23$}\\ \hline
	entropy/task+cm & 63.51 & \textcolor{red}{$81.29\pm0.37$} & \textcolor{red}{$84.94\pm0.1$} & \textcolor{red}{$85.82\pm0.86$} & \textcolor{red}{$87.44\pm0.17$} & \textcolor{red}{$88.65\pm0.68$} & \textcolor{red}{$89.56\pm0.43$} & \textcolor{red}{$90.22\pm0.48$} & \textcolor{red}{$90.85\pm0.45$} & \textcolor{red}{$91.03\pm0.35$} & \textcolor{red}{$91.39\pm0.35$}\\
	margin/task+cm & 63.51 & \textcolor{red}{$82.95\pm1.16$} & \textcolor{red}{$87.64\pm0.23$} & \textcolor{red}{$88.76\pm1.05$} & \textcolor{red}{$89.73\pm0.26$} & \textcolor{red}{$90.72\pm0.28$} & \textcolor{red}{$91.06\pm0.34$} & \textcolor{red}{$91.76\pm0.07$} & \textbf{\textcolor{blue}{$91.98\pm0.27$}} & \textcolor{red}{$92.52\pm0.22$} & \textcolor{red}{$92.72\pm0.16$}\\ \hline
	entropy/task+co+cm & 63.51 & \textcolor{red}{$80.98\pm1.46$} & \textcolor{red}{$84.94\pm0.68$} & \textcolor{red}{$85.86\pm0.19$} & \textcolor{red}{$87.86\pm0.43$} & \textcolor{red}{$88.78\pm0.16$} & \textcolor{red}{$89.36\pm0.2$} & \textcolor{red}{$90.47\pm0.52$} & \textcolor{red}{$90.88\pm0.29$} & \textcolor{red}{$91.29\pm0.41$} & \textcolor{red}{$91.95\pm0.13$}\\
	margin/task+co+cm & 63.51 & \textbf{\textcolor{blue}{$83.69\pm0.81$}} & \textbf{\textcolor{blue}{$87.86\pm0.45$}} & \textbf{\textcolor{blue}{$89.34\pm0.27$}} & \textcolor{red}{$89.91\pm0.05$} & \textcolor{red}{$90.8\pm0.28$} & \textbf{\textcolor{blue}{$91.72\pm0.24$}} & \textcolor{red}{$91.98\pm0.09$} & \textcolor{red}{$91.94\pm0.25$} & \textbf{\textcolor{blue}{$92.62\pm0.09$}} & \textbf{\textcolor{blue}{$92.81\pm0.19$}}\\ \hline
	maximum gap & 0 & 5.89 & \textbf{7.79} & 5.43 & 5.54 & 5.32 & 4.73 & 5.43 & 4.37 & 3.97 & 3.66\\
\hline
\end{tabular}
}
%\vspace{-5mm}
\end{table*}

% magnitude 얘기 뺐음
In Figure 2(b), there are examples of the \emph{directional} derivative of each loss for low uncertainty samples. The samples in left-hand side of the dotted line indicates low uncertainty cases.
Since samples with low uncertainty has a similar softmax output to the input sample even after data corruption, it's derivative has a similar direction to the target loss $\nabla L_{ce}$ proceeds.
% 여기부터는 원본이랑 의미가 달라서 원본 주석처리 남긴 것
In this situation, the direction of $\nabla L_{co}+\nabla L_{cm}$ is much like the input sample's own $\nabla L_{ce}$. As a result, the gradient of total loss $\nabla L_{total}$ has a similar effect to taking a larger step in a direction similar to the direction of inclination of $\nabla L_{ce}$ to proceed. In other words, regularization loss is less likely to diversify the information that the model attains.
This is likely to help with fast convergence similar to training with a large learning rate, but it is difficult to help find a solution with good generalization performance.
Conversely, for input samples with large uncertainties (examples to the right of the dashed line), the derivative of each loss is more likely to be give a different direction to the total loss. As a result, $\nabla L_{total}$ is likely to have a gradient fall in a direction different from the direction in which $\nabla L_{ce}$ proceeds, which means that there is a high possibility of performing a parameter update that is advantageous to finding a point potentially having high generalization performance. For this reason, it is presumed that training methodologies, including data augmentation-based uncertainty estimations and consistency-based regularization losses, improve generalization performance.
%In this situation, the absolute magnitude of $\nabla L_{co}+\nabla L_{cm}$ is generated as high as the similarity with the input sample. As a result, the gradient of total loss $\nabla L_{total}$ has a similar effect to taking a larger step in a direction similar to the direction of inclination of $\nabla L_{ce}$ to proceed. This is likely to help with fast convergence similar to training with a large learning rate, but it is difficult to help find a solution with good generalization performance. Conversely, for input samples with large uncertainties (examples to the right of the dashed line), the absolute magnitude of the derivative of each loss is greater than for the input samples with low uncertainties but is more likely to be given in a different direction than the target loss. As a result, $\nabla L_{total}$ is likely to have a gradient fall in a direction different from the direction in which $\nabla L_{ce}$ proceeds, which means that there is a high possibility of performing a parameter update that is advantageous to finding a point potentially having high generalization performance. For this reason, it is presumed that training methodologies, including data augmentation-based uncertainty estimations and consistency-based regularization losses, improve generalization performance.

% UPDATE FIG 4&5
% FIGURE3 : 여기로 옮긴 이유 - Table 1,2의 visualization이 figure3이라서.
\begin{figure}    %[t!]\
\centering
\includegraphics[width=1.0\linewidth]{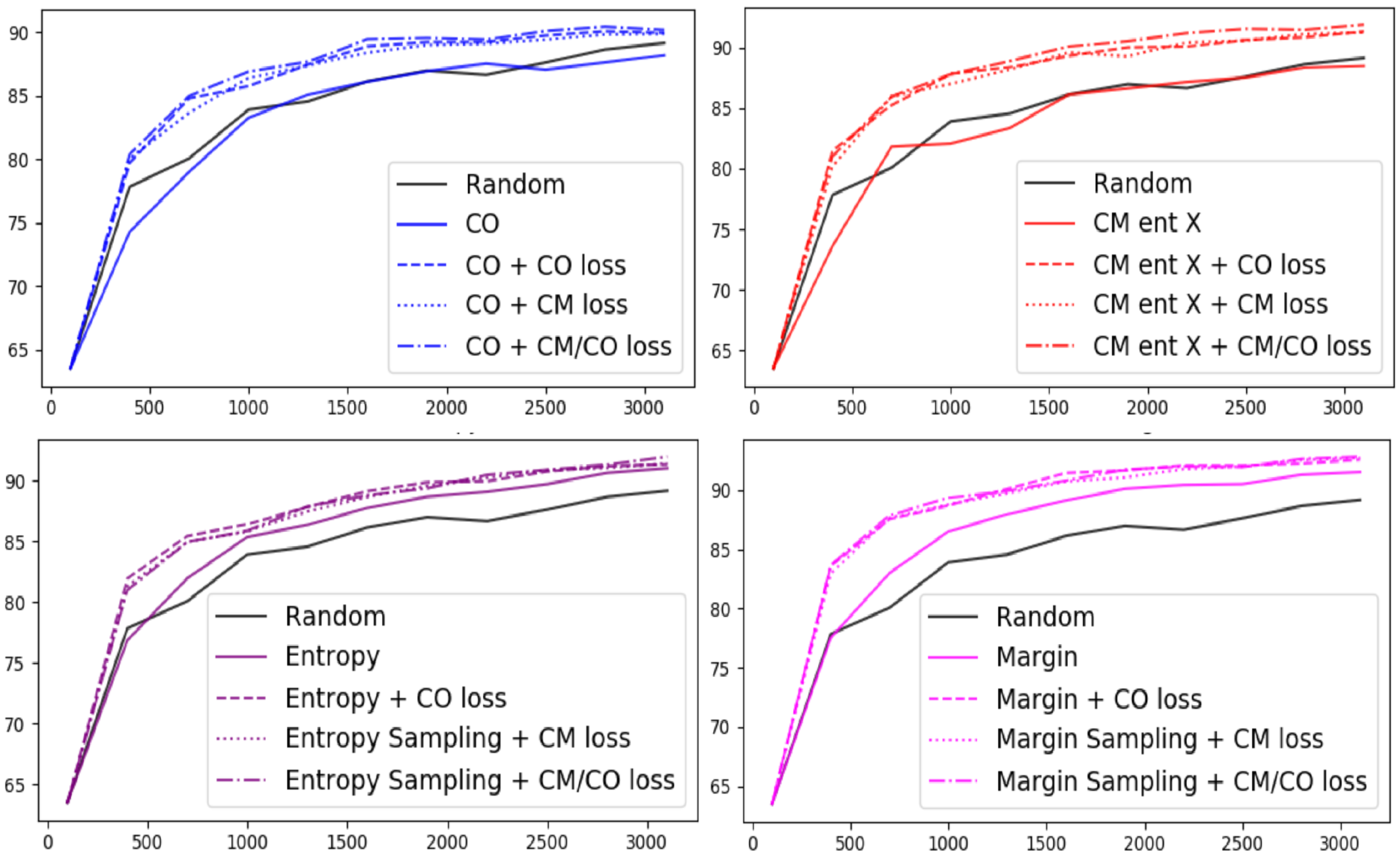}
%\vspace{-4mm}
\caption{\textbf{Active learning performance on FashionMNIST with different strategies and regularization losses.} For all graphs, the red line shows the performance of random sampling without any regularizer, (a) shows the performance of active learning using uncertainty sampling based on cutout, cutmix, entropy and margin in clockwise direction, and (b) shows the active learning performance using no loss, $L_{co}$, $L_{cm}$, and $L_{co}+L_{cm}$ based uncertainty sampling in clockwise direction, respectively.}
\label{fig:example}
\vspace{-4mm}
\end{figure}

%--------------------- EXPERIMENTS -------------------------------------------------

%\vspace{-1mm}
\section{Experiment Results}
%\vspace{-1mm}
% 말 중복이 많아서 제거
In order to verify the performance of active learning using augmentation-based methods, we conducted experiments with CNNs in various budget scenarios with multiple image classification datasets. For this purpose, we divided the experiments into three cases to analyze the effect of our proposed methods on active learning.
The first is whether data augmentation-based regularization can improve performance in active learning scenarios when \emph{uncertainty and regularization loss are utilized together}. For this, we tested whether the consistency-based regularization loss with arbitrary uncertainty measures can lead to performance improvement.
The second is a comparative analysis of the proposed active learning technique for \emph{different datasets}. By using representative datasets in an image classification problem, we confirmed that our methods can achieve consistent performance improvement in different datasets.
Finally, we conducted active learning experiments on \emph{multiple budget scenarios}. \cite{Simeoni19} showed that the results of active learning experiments using deep neural networks can be very sensitive to budget scenarios. We verified the robustness of our active learning methods on multiple budgets in the controlled experimental design.
For the overall performance report, please refer to Figure 4.\\

% 새로 이동된 FIGURE4. 이 그래프가 중요하기때문에 올렸음.
% FIGURE4
\begin{figure*}[t!]
%\vspace{-4mm}
\centering
\includegraphics[width=1.0\linewidth]{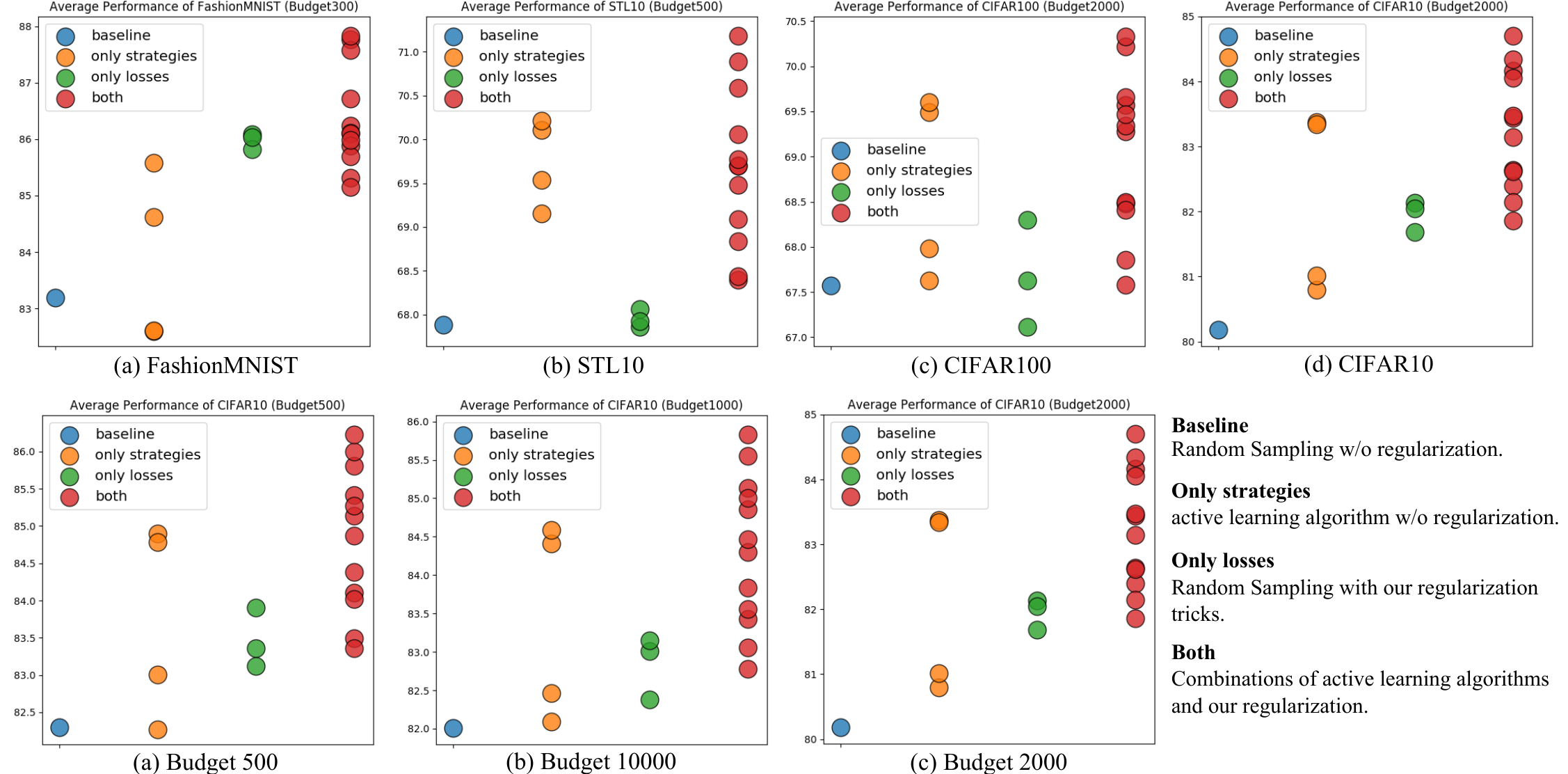}
\caption{\textbf{Active learning results for different datasets and budgets.} 
Each circle represents the average performance of entire epochs on the specific picking rule and regularization setting. It covers every possible combinations according to our active learning scenario. \textbf{(Exp A) The first row of the figure.} Average performance under different datasets. \textbf{(Exp B) The second row of the figure.} Average performance with CIFAR10 dataset under different budgets: 500, 1000, 2000. Although there are a difference in tendencies according to the budget and dataset, the results of using the consistency-based regularization losses with picking rule in all scenarios achieved the highest performance.
}
\label{fig:example}
\vspace{-5mm}
\end{figure*}

\vspace{-1mm}
\noindent \textbf{Dataset.} We conducted a series of experiments to validate multiple active learning scenarios of representative image recognition tasks. FashionMNIST \cite{Xiao17}, CIFAR10 \cite{Krizhevsky09}, CIFAR100 \cite{Krizhevsky09}, and STL10 \cite{Coates11} were used for the experiment. FashionMNIST consists of black and white images of 10 classes of clothing, each of which has a size of 28 by 28. The dataset consists of 60,000 training images and 10,000 test images. The CIFAR10 consists of 32 by 32 color images of 10 natural scenes, divided into 50,000 training images and 10,000 test images. CIFAR100 is designed to solve fined-grained image classification problems by subdividing data into 100 classes with detailed class information about CIFAR10. The STL10 is a dataset designed to solve the problem of classifying large amounts of test data with a relatively small number of labeled images, with ten classes for color images of size 96 by 96. Each class consists of 500 training samples and 800 test samples. \\

\vspace{-1mm}
\noindent \textbf{Training Detail.} As a data transformation to prevent overfitting during the training process, the random flip was performed after applying the 4-pixel padding that is common to all datasets. We use ResNet18 \cite{He16} as a training model for image classification. The initial labeled data $\mathcal{L}_{b,0}$ for active learning is fixed under the same configuration obtained with uniform samples, and initial model $h_{\mathcal{L}_{b,0}}(x,y;\theta_0)$ is set to the same weight initialization. In each $t+1$ active learning cycle, $b$ number of samples were selected after uncertainty measurements on the unlabeled sample set $\mathcal{U}_{N-b,t}$ to find label candidate samples using the current model $h_{\mathcal{L}_{b,t}}(x,y;\theta_t)$. After selecting the candidate samples, samples were added to $\mathcal{L}_{b,t}$ to perform training in $t+1$ cycles starting from $\mathcal{L}_{b,t+1}$ label samples with the same weight initial value as the model $h_{\mathcal{L}_{b,0}}(x,y;\theta_0)$. In all training, we use a stochastic gradient descent (SGD) optimizer, with an initial learning rate $0.1$, a momentum $0.9$, a weight decay $5e-4$, and a mini batch size $128$, and total $200$ epochs for each active learning cycle. The learning scheduler applied a scale of $0.1$ at $160$ epochs for all datasets except CIFAR100, and $0.2$ scales at $60$, $120$, and $160$ epochs for training using the CIFAR100 dataset.

%%%%%%%%%%%% FIGURE3 여기였음 %%%%%%%%%%%%%

% 번역체 문장 수정
% entropy & margin 사용한 이유 변경
%\vspace{-1mm}
\subsection{Deep active learning using data augmentation-based regularization}
%\vspace{-1mm}
To verify our suggesting active learning methods, we measured the performance according to the uncertainty measurement method. To check whether our methods can be extended to existing active learning uncertainty methods, we used entropy- and margin-based uncertainty measures with the same CNN model as the baseline learning method. Although a variety of uncertainty measures have been proposed \cite{Sener18} in addition to the two approaches, recent studies have shown sensitive results as changes in the dataset and budget scenarios occur \cite{Simeoni19}. Furthermore, in the case of \cite{Yoo19} that takes advantage of a submodule attached to the deep neural networks, it is difficult to accurately compare the performance because of additional parameters of the submodule.

The results of active learning using cutout and cutmix under a specific budget scenario and dataset are shown in Table 1 and Table 2. Since the proposed approach is divided into the data augmentation technique in the uncertainty measurement and regularization loss in the training process in the active learning scenario, the performance is divided into the sampling and learning strategy.
In Table 1, we can verify that the proposed active learning technique shows better performance than the entropy- and margin-based active learning techniques.
The performance variation of the proposed consistency-based regularization loss combined with other uncertainty measures is shown in Table 2.
In addition to the data augmentation-based uncertainty measurement method proposed here, it is evident that the method shows good performance in combination with arbitrary uncertainty measurement methods.

%----------------------------------
In Figure 3, graphical examples of performance variation with a particular budget scenario ($b=300$) in the FashionMNIST dataset are shown. In all graphs, the solid black line represents the result of training with target loss only on random sampling. The graphs in Figure 3 compare with several active learning scenarios with fixed uncertainty measurements. In the case of using the uncertainty technique without the consistency-based regularization loss, the performance of each active learning method is improved not that much. However, when the regularization loss was applied together during the training, it was confirmed that the performance steadily improved in all the uncertainty measurement-based active learning. In particular, the results of the bottom row show that the performance improvement is also shown for the uncertainty measurement method using entropy and margin, which means that other existing active learning methods can be combined with the regularization losses.

%---------- 그림이 변경되어 바뀐 부분 ----------

\subsection{Robustness for datasets}
To verify the effectiveness of the proposed active learning method in various datasets, image classification experiments were conducted on FashionMNIST, CIFAR10, CIFAR100, and STL10. In Figure 4 Exp A (first row), the average performances of the entire active learning cycle are visualized to show the performance gain efficiently. We divide the active learning scenarios into 4 approaches--'baseline' which is random sampling with no consistency-based regularization loss, 'only strategies' which is our proposed uncertainty method or existing rule with no regularization loss, 'only losses' which is random sampling with our regularization losses, and 'both' which is the combination of uncertainty methods with our regularization losses. Each colored-circle indicates the average accuracy in each setting. Our experiments are conducted among the following combinations: random sampling, cutout, cutmix, entropy sampling, and margin sampling as possible strategies, and no regularization, cutout, cutmix, and cutout with cutmix regularization as possible losses.

% 설명 다시 작성 - 엉성한 영어일 수 있음...!
% figure1과 연결짓고자함
First, Figure 4 Exp A visualize the average performance of each active learning settings in all dataset. It shows that consistency-based regularization loss combined with uncertainty measures consistently enhances the performance in all datasets.
Secondly, only the regularization loss or uncertainty measure alone may not effective due to the characteristics of each dataset, together exploits the natural synergy between their contribution that is explained in our schematic depiction in Figure 1. It is especially noteworthy that CIFAR100 and STL10, which have a relatively small number of images per class, do not show any significant performance improvement when only regularization loss is combined with random sampling, but show high performance in combination with uncertainty measurement.

\subsection{Robustness for budgets}
%\vspace{-1mm}
We conducted experiments based on the active learning scenarios with different budgets in a specific dataset to verify various budgets for the proposed active learning technique. For this purpose, experiments were conducted in $b\in\{500,1000,2000\}$ in the CIFAR10 dataset. The results of active learning under different budgets are shown in Figure 4 Exp B. Starting from a small amount of labeled data to a sufficient scale of selected data, we can see the performance gain on average. In other words, if a certain amount of label data is secured, there are overall performance increases in all cases especially when arbitrary data sampling budget rule is combined with our suggested data augmentation-based regularization loss. 

%------------------------------ CONCLUSION ---------------------------------
%\vspace{-1mm}
\section{Conclusions}
%\vspace{-1mm}
We proposed active learning methodologies using the augmentation-based consistency estimation derived from the analytical learning theory. Based on the analytical learning theory, it can be confirmed that the variation of function obtained from the observed data can have a substantial influence on the generalization error of the learning model.
% 이 문장은 없앴음. 그리고 아래에 설명 추가
%At the same time, it was confirmed that the consistency-based regularization technique showed a considerable performance difference according to the data augmentation method.
By adopting our augmentation-based consistency estimation methods on each active learning cycle, we can achieve consistent performance improvement and achieve a high-performance improvement in combination with the previously proposed uncertainty measurement methods. At the same time, we redefined the active learning scenario related to deep learning and visually explained how the proposed methods work on our definition.

% 여긴 괜찮은 듯
Nevertheless, there are limitations in experimenting only with the image classification problem. The proposed active learning methodology needs to be applied to various image recognition tasks, and additional experiments are needed to see if the same conclusion can be reached for large data. We believe that the analysis of data using analytical learning theory can be a good starting point for accessing various learning methodologies. In particular, it is expected that it can be effectively used for semi- or self-supervised learning that uses unlabeled data, and it can be useful for various learning methodologies that utilize unlabeled data.

{\small
\bibliographystyle{ieee_fullname}
\bibliography{egbib}
}

\end{document}